\documentclass[conference]{IEEEtran}
\IEEEoverridecommandlockouts
\usepackage{cite}
\usepackage{amsmath,amssymb,amsfonts}
\usepackage{algorithmic}
\usepackage{graphicx}
\usepackage{textcomp}
\usepackage{xcolor}
\usepackage{multirow} 
\usepackage{booktabs}
\def\BibTeX{{\rm B\kern-.05em{\sc i\kern-.025em b}\kern-.08em
    T\kern-.1667em\lower.7ex\hbox{E}\kern-.125emX}}
\begin{document}

\title{Interpretable Traffic Responsibility from Dashcam Video via Legal Multi-Agent Reasoning}

\author{
Jingchun Yang, Jinchang Zhang%
\thanks{Jingchun Yang is with Northeast University. Jinchang Zhang is with SUNY Binghamton University.}
}

\maketitle

\begin{abstract}
The widespread adoption of dashcams has made video evidence in traffic accidents increasingly abundant, yet transforming “what happened in the video” into “who is responsible under which legal provisions” still relies on human experts. Existing ego-view traffic accident studies mainly focus on perception and semantic understanding, while LLM-based legal methods are mostly built on textual case descriptions and rarely incorporate video evidence, leaving a clear gap between the two.  We first propose C-TRAIL, a multimodal legal dataset that, under the Chinese traffic regulation system, explicitly aligns dashcam videos and textual descriptions with a closed set of responsibility modes and their corresponding Chinese traffic statutes. On this basis, we introduce a two-stage framework: (1) a traffic accident understanding module that generate textual video descriptions; and (2) a legal multi-agent framework that output responsibility modes, statute sets, and complete judgment reports. Experimental results on C-TRAIL and MM-AU show that our method  outperforms general and legal LLMs, and existing agent-based approaches, while providing a transparent and interpretable legal reasoning process.
\end{abstract}

\begin{IEEEkeywords}
Multimodal Legal Reasoning, Legal Multi-Agent Systems, Chinese Traffic Law
\end{IEEEkeywords}

\section{Introduction}
With the widespread adoption of dashcams and other in-vehicle sensors, an increasing share of traffic accidents now have complete video records. Efficiently and reliably converting these videos, in realistic road settings, into responsibility attribution and concrete statutory provisions has become a key challenge for traffic police and the courts. On the one hand, the sheer scale and diversity of cases make manual analysis and statute lookup extremely costly; on the other hand, even with video evidence, legal professionals must still conduct multi-level multimodal reasoning—clarifying what happened, who is responsible, and which specific legal articles apply.

In the multimodal domain, extensive research has focused on perception and semantic understanding of ego-view traffic accident videos, including accident detection, classification, and accident reason answering.  These methods \cite{zang2023discovering}  can understand and describe traffic scenes and, in a QA form, answer “why did the accident happen?”, but their outputs mostly remain at the perceptual or semantic level—such as “whether an accident occurred” or “what type of accident it is”—and have not been elevated to the legal level of responsibility attribution and statute application.
Meanwhile, legal large language models \cite{hong2023metagpt,he2023lego} and multi-agent judge frameworks have shown promising potential for judicial reasoning and simulated court debates, but they are almost entirely built on a single modality of “case texts + statutes”, without incorporating video evidence into the legal reasoning process. In particular, no existing work has, in concrete traffic accident scenarios, established an explicit and interpretable mapping from observable behaviors to responsibility modes and statutory provisions.
As a result, there remains a clear gap between traffic video understanding and legal reasoning: video  research largely stops at accident-level semantic understanding, while legal methods mostly ignore  visual evidence. At present, there is no unified framework that, under the Chinese road traffic law system, can systematically map multimodal evidence (video and text) to responsibility modes and specific articles of law, while providing a traceable judicial decision process.
To bridge this gap, we propose an interpretable traffic responsibility analysis pipeline under the Chinese road traffic law framework, centered on dashcam videos. We first construct the multimodal legal dataset C-TRAIL, which explicitly aligns ego-view videos and accident texts with predefined responsibility modes and their associated statutes. Then, we design an accident understanding module that integrates ego-motion and vehicle state to automatically generate structured textual descriptions from raw videos. On this basis, we introduce a legal multi-agent framework that outputs responsibility modes, statute sets, and complete judgment-style reports.

\begin{figure*}[t]
\begin{center}
\includegraphics[width=17cm, height=4.4cm]{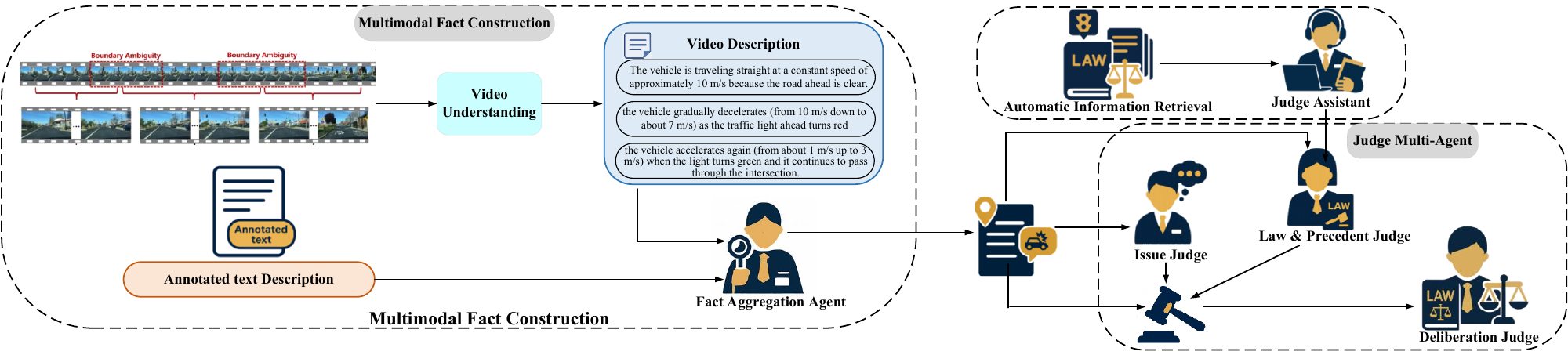}
\end{center}
\vspace{-6mm}
\caption{Overview of the judge multi-agent framework. On the left is the multimodal fact construction module: a video understanding module first segments the video into multiple events and generates event-level video descriptions, which are then combined with the accident text annotations provided in the dataset and integrated by the Fact Aggregation Agent into a unified case fact statement. The upper right shows the legal resource retrieval module: based on the fact statement, the Judge Assistant retrieves relevant statutory provisions and typical cases from the traffic law knowledge base and external resources. The lower right shows the judge multi-agent module: the Issue Judge analyzes the case facts and responsibility modes, the Law-Precedent Judge reviews and supplements the applicable statutes and precedents, and the Deliberation Judge consolidates these opinions to produce the final liability determination and judgment.
}
\vspace{-7mm}
\label{archagent}
\end{figure*}

\begin{figure*}[t]
\begin{center}
\includegraphics[width=15cm, height=4.5cm]{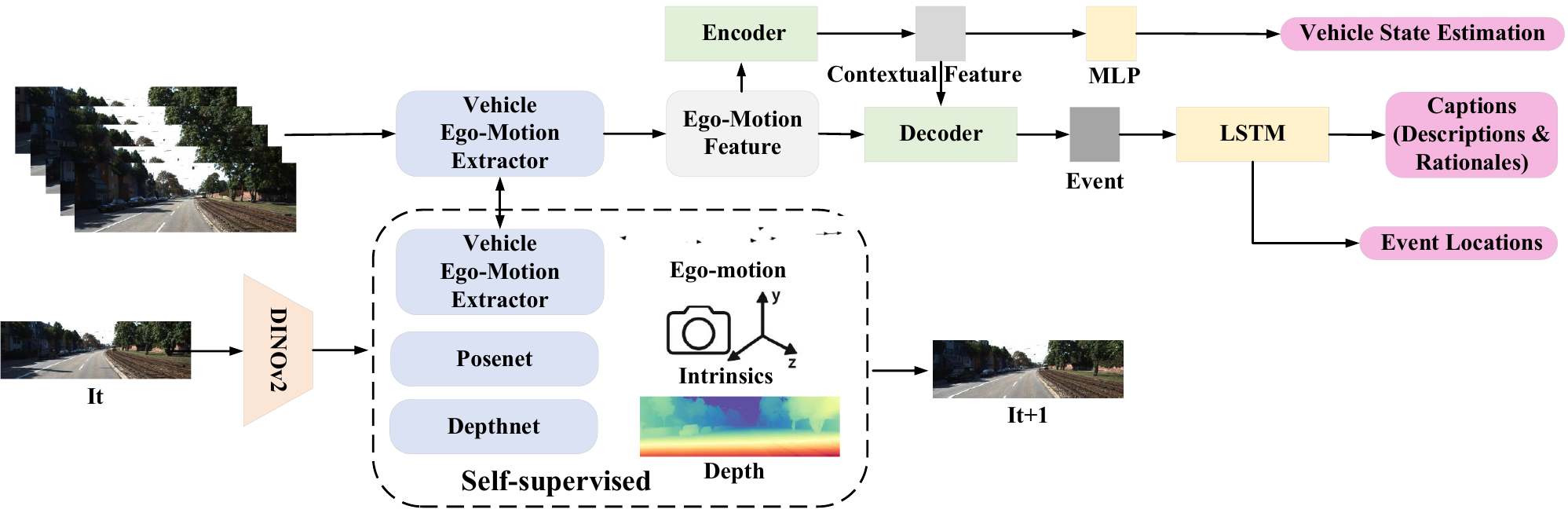}
\end{center}
\vspace{-7mm}
\caption{Overall architecture of video understanding and description framework.
Given an input video, a Vehicle Ego-Motion Extractor first produces frame-wise ego-motion features, which are fed into a Encoder to obtain contextual feature. On the one hand, an MLP head regresses the vehicle state (speed and steering) from these contextual features; on the other hand, a DETR-style Decoder outputs event feature, which are further processed by an LSTM to generate dense captions  together with their corresponding event locations. The Vehicle Ego-Motion Extractor itself is pre-trained in a self-supervised manner by using  DepthNet to estimate depth, PoseNet to estimate camera intrinsics, and photometric reconstruction between adjacent frames $I_t$ and $I_{t+1}$.
}
\vspace{-6mm}
\label{archego}
\end{figure*}

The main contributions of this paper can be summarized as follows:
1. The first multimodal responsibility analysis framework for Chinese traffic law. We propose a  framework of “video + text → responsibility mode → statute”. To the best of our knowledge, this is the first work that combines multimodal video–text inputs with a legal multi-agent adjudication framework and explicitly models the reasoning chain.
2. We propose C-TRAIL, the first multimodal legal dataset under the Chinese road traffic legal system that explicitly links dashcam videos and textual descriptions to responsibility modes and Chinese traffic law provisions.
3. We propose a traffic accident understanding framework as shown in Fig. \ref{archego}.
4.We design a legal multi-agent framework as shown in Fig. \ref{archagent}.

\section{RELATED WORK}
\subsection{Ego-view Traffic Accident Video Understanding}
Existing research on ego-view traffic accident videos mainly targets perception and semantic understanding of accidents, including detection, anticipation, classification, and reason answering. 
For accident classification, limited samples per category mean that work dedicated to ego-view accident classification is still sparse, and many approaches remain constrained to fixed surveillance viewpoints and low resolutions \cite{luo2023simulation}. ViT-TA uses attention maps to highlight critical objects and improve interpretability in accident classification \cite{kang2022vision}. 
For accident reason answering, 
SUTD-TrafficQA casts cause explanation and prevention advice as a QA task over dynamic traffic scenes \cite{xu2021sutd}, while later studies further explore cross-modal causal reasoning for traffic event understanding \cite{zang2023discovering}. 
Meanwhile, recent progress in efficient VLA modeling shows that fast-slow reasoning, instruction-driven routing, and semantic-aligned sparsification are effective for handling complex dynamic visual inputs under limited computational budgets \cite{li2025lion,li2025cogvla,li2025semanticvla}. 
These ideas are relevant to accident analysis because ego-view accident videos also require selective attention to legally salient objects, actions, and temporal evidence. Related studies have also explored integrating geometric priors, and VLM for more structured scene understanding, \cite{zhang2025vision,zhang2024embodiment}.
Nevertheless, existing QA-based methods still remain at the level of semantic explanation and do not explicitly support judicial decision-making, such as legal responsibility attribution or the selection of concrete legal actions.
\vspace{-2mm}
\subsection{LLM-driven Legal Reasoning}
\vspace{-1mm}
With the rapid development of large language models (LLMs), AI for law has advanced notably. GPT-3 can classify legal and deontic rules with limited data \cite{liga2023fine}, LLMs have been used to improve laypeople’s legal literacy via storytelling, and to summarize lengthy judicial opinions \cite{deroy2024applicability}. However, these works still fall short of real-world justice systems, especially in capturing institutional complexity and dynamic collegial deliberation—for instance, \cite{hamilton2023blind} trains nine models on opinions of individual supreme court justices but does not model their joint decision-making. Recently, LLM-based multi-agent systems have shown promise for complex tasks: collaboration and cognitive diversity among agents improve problem solving \cite{he2023lego}, while domain-specific systems such as Agent Hospital in healthcare achieve better outcomes than single models through dynamic interaction \cite{hong2023metagpt}. 
\vspace{-2mm}
\section{Ego-motion–Aware Accident Understanding}

\vspace{-1mm}
\subsection{Self-supervised Learning}
\label{subsec:self_supervised}

We first train DepthNet and PoseNet under a self-supervised depth estimation framework.  
Given two consecutive frames $I_t$ and $I_{t-1}$, DepthNet predicts a dense depth map
for the reference frame, denoted as $D_{t-1}$, while PoseNet estimates the 6-DoF
relative camera pose $T_{t-1\rightarrow t}$ between the two frames. The camera intrinsics
 as $K$.
With the predicted depth and pose, pixels from the previous frame $I_{t-1}$ can be
projected into the current viewpoint to reconstruct the target image $\hat I_t$:
$
    \hat I_t = I_{t-1}\big\langle \operatorname{proj}(D_{t-1}, T_{t-1\rightarrow t}, K)\big\rangle ,
$
where $\operatorname{proj}(\cdot)$ represents the operation that transforms pixel
coordinates using the estimated depth and pose into the target view and samples
the corresponding intensities via bilinear interpolation.

\vspace{-2mm}
\subsection{Vehicle Ego-motion Extractor (VEE)}
\vspace{-1mm}
\label{subsec:ego_motion_extractor}

The PoseNet obtained above learns to predict the geometric relative motion from
frame $t-1$ to $t$. However, for our downstream tasks we are more interested
in a representation of the {driving behavior or control intention} at a given
time step, rather than pure frame geometry.
To this end, we adopt a two-stage training strategy. In the second stage, we freeze
the trained DepthNet and re-initialize PoseNet. We now feed only a single
frame $I_t$ into the network and require it to predict the same type of ego-motion
quantity as in the first stage, {i.e.}, ``what ego motion should be generated
given the current observation''. During this stage, DepthNet serves as a fixed
geometric prior, while PoseNet is forced to extract high-level semantic cues from
the single-frame appearance that are correlated with the ego vehicle's
motion.
The new PoseNet no longer merely estimates
frame-to-frame pose, but acts as a network that produces a driving-decision-related
representation conditioned on the current observation. Its intermediate features
encode both the ego-motion state and the scene context that leads to such motion.
We refer to this new PoseNet together with the DepthNet as the
VEE, from which all subsequent frame-level
ego-motion features are derived.

\vspace{-2mm}
\subsection{Vehicle State Estimation (VSE)}
\label{subsec:vse}
\vspace{-2mm}
On top of the VEE, we further design a supervised
 VSE task to explicitly constrain the ego-motion
representation, so that it better reflects the true driving state (speed and steering).
Given a video $V$ of the ego vehicle with $N$ frames, we have $M$ recorded
vehicle states:
$
    \{(v_1, s_1), \dots, (v_{M-1}, s_{M-1}), (v_M, \_)\},
$
where $v_t$ denotes the speed at time step $t$, and $s_t$ denotes the steering
between time step $t$ and $t+1$. Since the vehicle states are typically sampled
at a fixed time interval rather than at every frame, we have $M < N$.
First, the $N$ video frames are passed through the Ego-motion Extractor and a
temporal Transformer encoder to obtain a sequence of contextual frame features
$
    f_1, f_2, \dots, f_N .
$
Two independent MLP heads are used to regress speed and steering, respectively.
The VSE loss as: 
\vspace{-2mm}
\begin{equation}
    L_{\mathrm{mse}}
    = \frac{1}{M}\sum_{i=1}^{M} (v_i - v'_i)^2
    + \frac{1}{M-1}\sum_{i=1}^{M-1} (s_i - s'_i)^2 .
    \vspace{-1mm}
\end{equation}

This module serves as an explicit regression task that encourages the learned
features to truly encode ``speed + steering''. At the same time,
since the VSE head shares the ego-motion representation with the main network,
its supervision signal also enhances the modeling of ego-motion
state for subsequent event representation and caption generation.
\vspace{-2mm}
\subsection{Event-level Representation and Captioning}
\label{subsec:event_repr_caption}

After obtaining the contextual frame-level feature sequence from the Vehicle Ego-motion Extractor and the temporal Transformer encoder, we denote the features of a video with $T$ frames as
$
\{f_1, f_2, \dots, f_T\},
$
where each $f_t$ encodes both visual appearance and ego-motion information.
We then introduce an event decoder $\mathrm{Dec}_{\mathrm{event}}$ that directly operates on the entire frame-level feature sequence and maps it into a set of event-level representations. Formally, we write
$
    \{e_1, e_2, \dots, e_N\} = \mathrm{Dec}_{\mathrm{event}}(\{f_t\}_{t=1}^T),
$
where $\mathrm{Dec}_{\mathrm{event}}(\cdot)$ denotes the event decoder, $e_n$ is the semantic feature of the $n$-th event, and $N$ is the number of predicted events (or a predefined maximum number of events). Intuitively, the event decoder aggregates the frame-level features along the temporal dimension and compresses a contiguous segment of frames into a single event-level vector $e_n$, such that each $e_n$ corresponds to a semantically coherent driving event.
Given the event features $\{e_n\}$, we further predict the temporal boundaries of each event on the video timeline. A temporal localization head $\mathrm{MLP}_{\mathrm{loc}}$ regresses the start and end positions of each event:
$
    [l^{(s)}_n,\, l^{(e)}_n] = \mathrm{MLP}_{\mathrm{loc}}(e_n),
$
where $l^{(s)}_n$ and $l^{(e)}_n$ denote the start and end of the $n$-th event, respectively, which can be represented as normalized frame indices or timestamps. Collecting the predictions for all events yields the event set
$
    \mathcal{E} = \{(e_n,\, l^{(s)}_n,\, l^{(e)}_n)\}_{n=1}^N,
$
which explicitly specifies ``which events occur in the video'' (through the set size and event features) and ``where each event happens in time'' (through its temporal boundaries).
On top of the event localization, we employ an LSTM-based text decoder to perform conditional caption generation for each event. Let the word sequence associated with the $n$-th event be
$
Y_n = (y_{n,1}, y_{n,2}, \dots, y_{n,L_n}),
$
then the generation process of the scene description and rationale for the $n$-th event can be written as
$
    Y_n = \mathrm{LSTM}_{\mathrm{dec}}(e_n),  n = 1,\dots,N,
$
where $\mathrm{LSTM}_{\mathrm{dec}}(\cdot)$ takes the event feature $e_n$ as the initial hidden state or conditioning vector and autoregressively produces the natural-language sequence, including what happens, which agents are involved, and why the situation leads to risk or collision.
During training, the temporal localization head $\mathrm{MLP}_{\mathrm{loc}}$ and the text decoder $\mathrm{LSTM}_{\mathrm{dec}}$ can be jointly optimized, encouraging the event features $e_n$ to reside in a shared semantic space that supports both boundary prediction and caption generation.

% As a result, for each ego-view driving video the model can explicitly output:
% (i) the event set $\mathcal{E}$ (the number of events and their semantic features);
% (ii) the temporal boundaries $[l^{(s)}_n,\, l^{(e)}_n]$ for each event; and
% (iii) the aligned scene descriptions and rationales $Y_n$.
% These structured event-level outputs provide a clear temporal foundation for subsequent accident causality analysis, liability attribution, and statute grounding.

\section{Multi-Agent Judicial}
\subsection{Multimodal Fact Construction Module}
\label{sec:fact-construction}

After generating video descriptions in the previous section, we obtain for each dashcam video a natural-language description \(T_v\), which covers vehicle motion, road environment, and the collision process. The MM-AU dataset  provides manually annotated text \(T_a\) that focuses more on accident causes and liability-related factors and is therefore more reliable; in contrast, \(T_v\) is closer to the raw visual content and supplies fine-grained scene and behavior details.
Based on this, we design a multimodal fact construction module that takes \((T_a, T_v)\) as input and uses a {Fact Aggregation Agent} to produce a unified case fact statement \(F\). The agent first extracts core case elements from the high-confidence text \(T_a\) (e.g., roles of the involved vehicles, road type, key driving behaviors, and collision type), and then parses \(T_v\) to mine additional details that may be useful for adjudication, performing a consistency check between the two sources: if the information in \(T_v\) is compatible with \(T_a\), it is incorporated to enrich the fact set; if they conflict on core facts such as direction of travel, key maneuvers, or the potentially liable party, the agent gives priority to \(T_a\) and downweights or discards the conflicting information from \(T_v\).
Through this confidence-aware fusion strategy, the module produces a final fact statement \(F\) that preserves the reliability of \(T_a\) in terms of causal and liability-related factors while, as far as possible, incorporating the fine-grained environmental and behavioral information provided by \(T_v\), thereby mitigating the impact of automatic captioning errors on the overall factual assessment. Legal retrieval and judgment refinement modules all take \(F\) as their input.

\vspace{-2mm}
\subsection{Judge Assistant}
After completing the multimodal fact construction, we obtain a unified fact statement 
F for the current case. The Judge Agent is equipped with general legal knowledge and reasoning abilities, but is not specifically specialized in the traffic law domain. To enhance its expertise and timeliness in traffic accident scenarios, we introduce a Legal Resources Retrieval module, in which a retrieval-capable Judge Assistant agent provides external support focused on traffic regulations and representative cases.
\textbf{Judge Assistant.}
% We instantiate a dedicated agent as the Judge Assistant to help the Judge Agent, who already possesses general legal literacy, access more fine-grained and traffic-specific external legal resources. Given the fact statement 
% F as input, the Judge Assistant (i) retrieves relevant statutory provisions and typical traffic accident cases from our constructed traffic law knowledge base, and (ii) can also access publicly available online resources, such as accident bulletins issued by traffic authorities, authoritative examples of liability determinations in similar accidents, and public-facing explanations of traffic regulations.
% These materials do not directly replace the judgment of the Judge Agent; rather, they serve as supplementary evidence and domain-specific background, helping the Judge Agent, on top of its inherent legal reasoning ability, to produce decisions that better align with real-world traffic law practice. The Judge Assistant filters, summarizes, and structures the retrieved documents, compressing the relevant statutory articles, key points of precedents, and normative explanations into a concise legal information summary, so that the Judge Agent can use them efficiently within a limited context window, balancing information richness with controllability of the reasoning process.
We construct a dedicated “judge assistant” agent to provide the judge agent, which already possesses general legal competence, with finer-grained, traffic-specific external legal support. Taking the fact description 
F as input, the judge assistant on the one hand retrieves, from our traffic law knowledge base, legal provisions and representative traffic accident cases related to the current accident pattern; on the other hand, it can also access publicly available online resources, such as accident bulletins issued by traffic authorities, exemplary liability determinations from authoritative institutions, and public-oriented interpretations of traffic regulations.
This information does not directly replace the judgment made by the judge agent; instead, it serves as supplementary evidence and domain background, improving the consistency of its decisions with real-world traffic law practice. The judge assistant filters, summarizes, and structurally organizes the retrieved documents, compressing the relevant legal provisions, key precedents, and normative explanations into a concise “legal information summary.”
\textbf{Automatic Information Retrieval.}
% For knowledge base retrieval, our traffic law knowledge base includes the Road Traffic Safety Law and its supporting regulations, related judicial interpretations, typical judicial decisions, and traffic accident liability determination documents. Based on the core elements contained in F (e.g., accident type, collision pattern, road type, and behavior patterns of each party), the Judge Assistant first infers the accident category to which the current case belongs, and then uses the BM25 model \cite{yates2021pretrained} to perform coarse retrieval, selecting the top 100 candidate documents from the knowledge base.
% Subsequently, the Judge Assistant applies the BGE-Large model \cite{bge_embedding} to encode these candidate documents into vectors and perform semantic re-ranking, selecting the document that is most similar to the current case in terms of factual pattern and legal issues as the optimal precedent.
% To provide sufficient legal grounding while controlling context length, the Judge Assistant further extracts explicitly cited traffic law provisions from the top 10 re-ranked precedents, aggregating them into a set of candidate statutory articles closely related to the current case. Finally, the Judge Assistant delivers both the summary of the optimal precedent and the assembled article set to the Judge Agent, which then uses them in the subsequent judgment generation and refinement stages.
In terms of knowledge-base retrieval, we construct a traffic law knowledge base that covers the Road Traffic Safety Law and its supporting regulations, relevant judicial interpretations, representative judicial decisions, and traffic accident liability determination documents. Based on the key elements in the fact description 
F, the judge assistant first infers the accident category of the case and then performs a coarse retrieval over the knowledge base using the BM25 model \cite{yates2021pretrained} to obtain the top 100 candidate documents. Next, the judge assistant applies the BGE-Large model \cite{bge_embedding} to encode these candidates into vectors and conduct semantic re-ranking, selecting the top 10 documents that are most similar to the current case in terms of fact patterns and legal issues, and aggregating them into a candidate statute set. Finally, the judge assistant provides the organized analysis to the judge agent for subsequent judgment generation and refinement.
\vspace{-2mm}
\subsection{Judge Multi-Agent }
In this module, we do not delegate the entire judgment process to a single judge agent. Instead, inspired by the collegial panel mechanism in real-world judicial practice, we decompose the ``judge'' role into multiple sub-agents with different emphases. These sub-agents analyze the case independently on the same factual basis and provide mutual checks and balances. We construct a judge multi-agent framework consisting of three types of judge agents: an {Issue Judge}, a {Law \& Precedent Judge}, and a {Deliberation Judge}.
The {Issue Judge} takes the unified fact statement \(F\) as input and, relying only on its inherent general legal knowledge, identifies and decomposes the core legal issues in the case Based on this analysis, it produces a \textit{preliminary judgment}, including a liability allocation, a set of preliminarily applicable traffic law provisions, and the corresponding reasoning.
The {Law-Precedent Judge} also operates on the fact statement \(F\), but additionally receives the legal information summary produced by the Legal Resources Retrieval module (including the candidate statutory provisions and the optimal precedent summary). Its focus is not on re-evaluating the facts, but on reviewing the preliminary judgment from the perspective of consistency between statutes and precedents: it examines whether the applied provisions are accurate, whether there are omissions or over-extensions, and, by referring to liability determinations and sentencing patterns in similar cases, proposes \textit{revision suggestions} regarding both liability allocation and the choice of legal provisions. 
The {Deliberation Judge} simultaneously receives the fact statement \(F\), the preliminary judgment, and the revision suggestions, and serves as the ``final decision-maker'' in a virtual collegial deliberation. It synthesizes and balances the views of the different judge sub-agents: on the one hand, preserving the basic structure of the initial analysis; on the other hand, fully incorporating feedback from the statutory and precedent dimensions. Ultimately, the Deliberation Judge produces the \textit{final judgment}, including the final liability label, the final set of applicable provisions, and a structured case analysis and reasoning section that clearly presents the reasoning chain from ``facts'' to ``judgment''.
In implementation, all three judge agents are instantiated from the same underlying legal LLM, and functional differentiation is achieved solely through distinct role prompts and instruction constraints.

% This design of ``shared foundational capability with multiple judicial perspectives'' not only fully exploits the model's general legal knowledge, but also mitigates the bias and instability that may arise when a single model handles complex judgment tasks, via explicit role division and multi-round interaction.

\vspace{-3mm}
\section{Experiments}
\vspace{-3mm}
\subsection{Dataset}
\textbf{C-TRAIL.}
We construct the C-TRAIL dataset (Traffic Responsibility Analysis from an Interpretable Chinese Legal Perspective), which explicitly links multimodal evidence (video and text) with legal responsibility and concrete traffic statutes under the framework of Chinese road traffic law.
We first define a closed set of core responsibility modes and their associated statute sets within the Chinese road traffic legal system. These modes cover high-frequency liability patterns in real-world accidents that can be clearly identified from video. 
% such as rear-end collisions and insufficient following distance, unsafe lane changes/merging, violations of intersection right-of-way and yielding obligations, traffic signal and road marking violations, failure to adequately protect pedestrians and non-motorized road users, and wrong-way driving or improper lane usage.
Each responsibility mode is pre-bound to a set of relevant Chinese traffic law provisions, with detailed definitions and statute sets provided in the appendix.
Building on the MM-AU dataset \cite{fang2024abductive}, we tailor the data to Chinese traffic scenarios. We first use accident descriptions and annotated texts as input to a large language model to semantically filter out samples that are clearly incompatible with Chinese road environments. Then, a light manual review further removes cases that cannot be reasonably interpreted under the Chinese legal framework, yielding a refined subset of about 1,000 videos.
For annotation, we adopt a hybrid “automatic + manual” pipeline. A large language model first generates candidate responsibility modes and brief responsibility explanations based on each video and its texts. Annotators with driving experience and basic legal knowledge then check and revise these candidates to determine the final responsibility mode label. According to the predefined mapping between responsibility modes and statute sets, the corresponding statutes are automatically assigned for each case. 
\textbf{Evaluation metric.}
% The core task on C-TRAIL is responsibility mode classification: given the video, text, and the unified fact statement produced by our multimodal fact construction module, the model must select one label from the predefined responsibility mode set (RM1–RM6). We report two types of metrics:
% \textbf{Responsibility-level metrics.}
% We use Accuracy and Macro-F1 to evaluate classification performance, where Macro-F1 alleviates the impact of class imbalance across different responsibility modes. 
% \textbf{Legal-level metric — Core Statute Hit.}
% For each responsibility mode, we predefine 1–2 expert-selected core statutes within its statute set. Given the model’s predicted responsibility mode, we obtain the corresponding predicted statute set via the mode–statute mapping, and check whether it contains at least one core statute of the ground-truth mode. The resulting Core Statute Hit rate measures whether the model not only predicts the correct type of responsibility, but also captures the most critical legal basis associated with that mode.
C-TRAIL: Given the video, text, and the unified fact statement produced by the multimodal fact construction module, the model must select one label from the predefined responsibility mode set (RM1–RM6).
We use two types of evaluation metrics:
Responsibility-level metrics. Accuracy and Macro-F1 are used to assess classification performance.
Legal-level metric (Core Statute Hit). For each responsibility mode, 1–2 core statutes are predefined. Given the model’s predicted responsibility mode, we obtain its statute set via the mode–statute mapping and check whether it contains any core statute of the ground-truth mode.

\textbf{MM-AU Dataset.}
MM-AU \cite{fang2024abductive} contains 11,727 egocentric driving videos and over 2 million frames, with rich annotations including accident windows, precise accident timestamps, and manually written textual descriptions.
In this dataset, accident cause understanding is formulated as the Accident Reason Answering (ArA) task: after watching the video and reading the given text, the model must select the most appropriate accident reason from multiple candidate answers.
\vspace{-3mm}
\subsection{Comparison with Multimodal VideoQA Models}
\vspace{-1mm}
\begin{table}[t]
\centering
\begin{tabular}{lccc}
\hline
\multirow{2}{*}{Methods} & \multicolumn{2}{c}{MM-AU} & C-TRAIL \\
\cline{2-3}\cline{4-4}
 & Acc. (val.) & Acc. (test.) & Acc. \\
\hline
VGT~\cite{xiao2022video}          & 68.40 & 68.66 &  26.8    \\
FrozenGQA~\cite{xiao2024can} & 77.10 & 77.01 & 33.5 \\
CoVGT~\cite{xiao2023contrastive}      & 81.70 & 79.97 & 37.2     \\
SeViLA~\cite{yu2023self}    & {89.26} & {89.02} &39.7 \\
Ours    & \textbf{94.35} & \textbf{93.97} & \textbf{86.4}\\
\hline
\end{tabular}
\caption{Comparison with VideoQA models on MM-AU \cite{fang2024abductive} and C-TRAIL.}
\vspace{-11mm}
\label{mmau}
\end{table}

From Table \ref{mmau}, we can see that on the MM-AU dataset our method significantly outperforms existing multimodal video QA models on the ArA task. The key difference is that, instead of directly “picking an answer” from the candidates, we first use the judge multi-agent framework to generate a structured case report that includes the accident process, key behaviors, causal chain, and the final decision with the corresponding violated statutes, and then select the most appropriate option from the candidates based on this report. Since the report has already carried out explicit reasoning about “why the accident occurred” and “how responsibility should be allocated and which statutes apply,” our approach improves both accuracy and interpretability, and also makes it easier to trace errors when they occur.
On the C-TRAIL dataset, the task is upgraded to inferring responsibility types, which requires the model to explicitly model how legal responsibility is assigned. The compared methods degrade significantly in this setting, indicating that they are more inclined toward visual–text semantic matching and lack systematic reasoning along the full “behavior → responsibility type → statute” chain, often relying only on superficial similarity between text and statute options. In contrast, our framework produces interpretable judicial analyses, making responsibility attribution more consistent with legal logic and more explainable.
\vspace{-2mm}
\subsection{Legal Responsibility Prediction with LLMs and Agents}

\begin{table}[t]
\centering
\resizebox{0.47\textwidth}{!}{
\begin{tabular}{l l c c c}
\toprule
Category & Method & Acc (\%) & Macro-F1 (\%) & Core (\%) \\
\midrule
General LLMs & GPT-3.5      &79.6 & 68.7&61.3   \\
 & GPT-4       &82.9  &74.7  & 67.5 \\
 & DeepSeek-R1  &81.3  &73.2  &63.8 \\ \hline
 Legal LLMs   & LaWGPT \cite{zhou2024lawgpt}      & 81.8 & 72.4  & 64.1 \\
  & ChatLaw2-MoE \cite{cui2023chatlaw}   & 84.1 &75.3 &68.5  \\
   & Qwen2.5-Law & 84.7 &76.7  & 69.7 \\ \hline
Agent Methods & ReAct \cite{yao2022react}       &78.4  & 67.5 & 59.5 \\
 & AgentsCourt \cite{he2024agentscourt}&85.1  & 77.3 & 71.2 \\
& Ours        & \textbf{86.4}  & \textbf{79.1} & \textbf{73.2 } \\
\bottomrule
\end{tabular}}
\caption{Overall performance of our framework and baselines on C-TRAIL in terms of accuracy, macro-F1, and core statute hit.}
\vspace{-7mm}
\label{tab:ctrail_llm_comparison}
\end{table}

On the C-TRAIL dataset, all methods take as input a unified factual description formed by concatenating video-derived text with accident annotations.
As shown in Table~\ref{tab:ctrail_llm_comparison}, our multi-agent framework achieves the best results on Acc, Macro-F1, and Core statute hit rate, clearly surpassing general LLMs, legal LLMs, and existing agent-based systems.
General-purpose LLMs exhibit some ability in responsibility mode classification, but their Core scores are limited by the lack of explicit modeling of Chinese traffic regulations. Legal LLMs further improve Acc and Macro-F1, yet still mainly rely on semantic matching between descriptions and statute options. In contrast, our method explicitly models the reasoning chain {behavior}, {responsibility mode}, {statutes} via multimodal fact construction, judge-assistant–based legal retrieval, and a judge multi-agent deliberation mechanism, thereby improving overall accuracy and significantly increasing key statute hit rate, with a stronger grasp of legal responsibility and better interpretability.
\vspace{-2mm}
\subsection{Analysis of the Video Understanding Module}
\begin{table}[t]
\centering
\resizebox{0.47\textwidth}{!}{
\begin{tabular}{lllllll}
\multicolumn{1}{c}{Ego} & 
\multicolumn{1}{c}{VSE} & 
\multicolumn{1}{c}{Event} & 
\multicolumn{1}{c}{ Acc (\%) $\uparrow$  } & 
\multicolumn{1}{c}{Macro-F1 (\%) $\uparrow$ } & 
\multicolumn{1}{c}{Core (\%) $\uparrow$} & 
\\ \hline 
  &  &   &76.3  & 66.8 &57.2    
\\  \hline 
\checkmark    &  &    &84.8 &76.9   & 69.9
\\  \hline 
\checkmark   &\checkmark &     & 85.4 & 77.9&  72.6
\\  \hline  
\checkmark    & \checkmark &\checkmark   & \textbf{86.4}  & \textbf{79.1 }&\textbf{73.2} 
\\  \hline  
\end{tabular}}
\caption{Analysis Video Understanding module on the  C-TRAIL  dataset. Ego: Ego-Motion
Extractor. Event:  Event Locations.}
\vspace{-11mm}
\label{tab:ego-ablation}
\end{table}

As shown in Table \ref{tab:ego-ablation}, all three components of the video understanding module have a clear impact on downstream legal tasks, with the Ego-Motion Extractor being the most crucial foundation. Without any motion modeling and relying only on raw appearance features, the metrics are relatively low. Once ego-motion modeling (Ego) is introduced, all metrics increase significantly, indicating that self-vehicle behaviors such as deceleration, steering, and lane changes are critical for responsibility attribution and cannot be reliably captured from appearance frames alone.
Adding the VSE module to explicitly regress speed and steering further makes the motion representation closer to real driving states, and the results show that such vehicle-state supervision helps align responsibility modes and statutes more stably. Finally, incorporating the Event localization module, which segments the video into a series of “event segments,” allows the model to focus on key moments instead of modeling the entire long video uniformly. The additional performance gains demonstrate that event-level modeling can more effectively capture the crucial behaviors around the accident, thus providing more compact and targeted factual inputs for subsequent legal reasoning.
\vspace{-2mm}
\subsection{Ablation Study}
\begin{table}[t]
\centering
\resizebox{0.47\textwidth}{!}{
\begin{tabular}{llllllll}
\multicolumn{1}{c}{Caption} & 
\multicolumn{1}{c}{Fact Agg} & 
\multicolumn{1}{c}{Assistant} & 
\multicolumn{1}{c}{Mul Judge} & 
\multicolumn{1}{c}{ Acc (\%) $\uparrow$  } & 
\multicolumn{1}{c}{Macro-F1 (\%) $\uparrow$ } & 
\multicolumn{1}{c}{Core (\%) $\uparrow$} & 
\\ \hline 
  &  & &  &81.4  & 73.4 &63.7    
\\  \hline 
\checkmark    &  &  &  &82.3 &74.4   & 66.7
\\  \hline 
\checkmark   &\checkmark &   &  & 82.8 & 74.6&  67.3
\\  \hline  
\checkmark   &\checkmark & \checkmark &  &84.2  & 75.5 &68.8  
\\  \hline  
\checkmark    & \checkmark &\checkmark  & \checkmark  & \textbf{86.4}  & \textbf{79.1 }&\textbf{73.2} 
\\  \hline  
\end{tabular}}
\caption{Ablation study on the  C-TRAIL  dataset. Caption: Video Caption Module. Fact Agg: Fact Aggregation Agent. Assistant: Judge Assistant. Mul Judge: Judge Multi-Agent}
\vspace{-10mm}
\label{tab:overall-ablation}
\end{table}

Table~\ref{tab:overall-ablation} presents the ablation results on C-TRAIL. Using only the accident text as input (1 row), the metrics already indicate that the raw textual descriptions themselves contain strong signals for responsibility determination. Adding video-derived textual descriptions on this basis (2 row) brings a small improvement, showing that motion and interaction details from the video side provide additional cues, although simple concatenation is still insufficient to fully exploit the complementarity between the two text sources.
Further, introducing the Fact Aggregation Agent to rewrite video descriptions and accident texts into a legal fact statement (3 row) improves the result, suggesting that transforming evidence into a form closer to legal expression helps more stably align responsibility modes with statute sets. On top of this, enabling the Judge Assistant for external legal resource retrieval (4 row) leads to gains in both Acc and Core, indicating that professional regulations and cases are crucial for fine-grained statute application. Finally, under the same factual and legal inputs, extending a single judge to a multi-agent system (last row) yields the best overall result, confirming that introducing role specialization and mutual cross-checking on a shared fact base provides more reliable legal support for the final judgment.

\vspace{-3mm}
\section{Conclusion}
\vspace{-2mm}
Under the framework of Chinese road traffic law, this paper proposes an interpretable responsibility analysis system that links multimodal inputs to observable behaviors, responsibility modes, and statutory provisions. The system comprises three parts: the C-TRAIL multimodal legal dataset, an ego-motion–aware traffic accident understanding framework, and a legal multi-agent adjudication framework. Experiments show that our method surpasses general-purpose LLMs, legal LLMs, and existing agent-based approaches on responsibility mode prediction, core statute retrieval, and accident reason analysis, while producing clear and traceable judgments.

\bibliographystyle{IEEEbib}
\bibliography{icme2026references}

\begin{thebibliography}{10}

\bibitem{zang2023discovering}
Chuanqi Zang, Hanqing Wang, Mingtao Pei, and Wei Liang,
\newblock ``Discovering the real association: Multimodal causal reasoning in video question answering,''
\newblock in {\em CVPR}, 2023.

\bibitem{hong2023metagpt}
Sirui Hong, Mingchen Zhuge, Jonathan Chen, Xiawu Zheng, Yuheng Cheng, Jinlin Wang, Ceyao Zhang, Zili Wang, Steven Ka~Shing Yau, Zijuan Lin, et~al.,
\newblock ``Metagpt: Meta programming for a multi-agent collaborative framework,''
\newblock in {\em ICLR}.

\bibitem{he2023lego}
Zhitao He, Pengfei Cao, Yubo Chen, Kang Liu, Ruopeng Li, Mengshu Sun, and Jun Zhao,
\newblock ``Lego: A multi-agent collaborative framework with role-playing and iterative feedback for causality explanation generation,''
\newblock in {\em Findings of EMNLP 2023}, 2023.

\bibitem{luo2023simulation}
Haohan Luo and Feng Wang,
\newblock ``A simulation-based framework for urban traffic accident detection,''
\newblock in {\em ICASSP}. IEEE, 2023.

\bibitem{kang2022vision}
Minhee Kang, Wooseop Lee, Keeyeon Hwang, and Young Yoon,
\newblock ``Vision transformer for detecting critical situations and extracting functional scenario for automated vehicle safety assessment,''
\newblock {\em Sustainability}, 2022.

\bibitem{xu2021sutd}
Li~Xu, He~Huang, and Jun Liu,
\newblock ``Sutd-trafficqa: A question answering benchmark and an efficient network for video reasoning over traffic events,''
\newblock in {\em CVPR}, 2021.

\bibitem{li2025lion}
Wei Li, Bing Hu, Rui Shao, Leyang Shen, and Liqiang Nie,
\newblock ``Lion-fs: Fast \& slow video-language thinker as online video assistant,''
\newblock in {\em CVPR}, 2025, pp. 3240--3251.

\bibitem{li2025cogvla}
Wei Li, Renshan Zhang, Rui Shao, Jie He, and Liqiang Nie,
\newblock ``Cogvla: Cognition-aligned vision-language-action model via instruction-driven routing \& sparsification,''
\newblock in {\em NeurIPS}, 2025.

\bibitem{li2025semanticvla}
Wei Li, Renshan Zhang, Rui Shao, Zhijian Fang, Kaiwen Zhou, Zhuotao Tian, and Liqiang Nie,
\newblock ``Semanticvla: Semantic-aligned sparsification and enhancement for efficient robotic manipulation,''
\newblock in {\em AAAI}, 2026.

\bibitem{zhang2025vision}
Jinchang Zhang and Guoyu Lu,
\newblock ``Vision-language embodiment for monocular depth estimation,''
\newblock in {\em CVPR}, 2025.

\bibitem{zhang2024embodiment}
Jinchang Zhang, Praveen~Kumar Reddy, Xue-Iuan Wong, Yiannis Aloimonos, and Guoyu Lu,
\newblock ``Embodiment: Self-supervised depth estimation based on camera models,''
\newblock in {\em IROS}. IEEE, 2024.

\bibitem{liga2023fine}
Davide Liga and Livio Robaldo,
\newblock ``Fine-tuning gpt-3 for legal rule classification,''
\newblock {\em Computer Law \& Security Review}, 2023.

\bibitem{deroy2024applicability}
Aniket Deroy, Kripabandhu Ghosh, and Saptarshi Ghosh,
\newblock ``Applicability of large language models and generative models for legal case judgement summarization,''
\newblock {\em Artificial Intelligence and Law}, 2024.

\bibitem{hamilton2023blind}
Sil Hamilton,
\newblock ``Blind judgement: Agent-based supreme court modelling with gpt,''
\newblock {\em arXiv preprint}, 2023.

\bibitem{yates2021pretrained}
Andrew Yates, Rodrigo Nogueira, and Jimmy Lin,
\newblock ``Pretrained transformers for text ranking: Bert and beyond,''
\newblock in {\em Proceedings of the 14th ACM International Conference on web search and data mining}, 2021.

\bibitem{bge_embedding}
Shitao Xiao, Zheng Liu, Peitian Zhang, and Niklas Muennighoff,
\newblock ``C-pack: Packaged resources to advance general chinese embedding,'' 2023.

\bibitem{fang2024abductive}
Jianwu Fang, Lei-lei Li, Junfei Zhou, Junbin Xiao, Hongkai Yu, Chen Lv, Jianru Xue, and Tat-Seng Chua,
\newblock ``Abductive ego-view accident video understanding for safe driving perception,''
\newblock in {\em CVPR}, 2024.

\bibitem{xiao2022video}
Junbin Xiao, Pan Zhou, Tat-Seng Chua, and Shuicheng Yan,
\newblock ``Video graph transformer for video question answering,''
\newblock in {\em ECCV}, 2022.

\bibitem{xiao2024can}
Junbin Xiao, Angela Yao, Yicong Li, and Tat-Seng Chua,
\newblock ``Can i trust your answer? visually grounded video question answering,''
\newblock in {\em CVPR}, 2024.

\bibitem{xiao2023contrastive}
Junbin Xiao, Pan Zhou, Angela Yao, Yicong Li, Richang Hong, Shuicheng Yan, and Tat-Seng Chua,
\newblock ``Contrastive video question answering via video graph transformer,''
\newblock {\em TPAMI}, 2023.

\bibitem{yu2023self}
Shoubin Yu, Jaemin Cho, Prateek Yadav, and Mohit Bansal,
\newblock ``Self-chained image-language model for video localization and question answering,''
\newblock {\em NeurIPS}, 2023.

\bibitem{zhou2024lawgpt}
Zhi Zhou, Jiang-Xin Shi, Peng-Xiao Song, Xiao-Wen Yang, Yi-Xuan Jin, Lan-Zhe Guo, and Yu-Feng Li,
\newblock ``Lawgpt: A chinese legal knowledge-enhanced large language model,''
\newblock {\em arXiv preprint}, 2024.

\bibitem{cui2023chatlaw}
Jiaxi Cui, Munan Ning, Zongjian Li, Bohua Chen, Yang Yan, Hao Li, Bin Ling, Yonghong Tian, and Li~Yuan,
\newblock ``Chatlaw: A multi-agent collaborative legal assistant with knowledge graph enhanced mixture-of-experts large language model,''
\newblock {\em arXiv preprint}, 2023.

\bibitem{yao2022react}
Shunyu Yao, Jeffrey Zhao, Dian Yu, Nan Du, Izhak Shafran, Karthik~R Narasimhan, and Yuan Cao,
\newblock ``React: Synergizing reasoning and acting in language models,''
\newblock in {\em ICLR}, 2022.

\bibitem{he2024agentscourt}
Zhitao He, Pengfei Cao, Chenhao Wang, Zhuoran Jin, Yubo Chen, Jiexin Xu, Huaijun Li, Kang Liu, and Jun Zhao,
\newblock ``Agentscourt: Building judicial decision-making agents with court debate simulation and legal knowledge augmentation,''
\newblock in {\em Findings of EMNLP}, 2024.

\end{thebibliography}

\end{document}